\documentclass[11pt]{article}

\usepackage[utf8]{inputenc}
\usepackage[T1]{fontenc}
\usepackage{lmodern}
\usepackage[letterpaper,margin=1in]{geometry}
\usepackage{amsmath,amssymb,amsthm,mathtools}
\usepackage{graphicx}
\usepackage{booktabs,tabularx,array}
\usepackage{xcolor}
\usepackage[numbers,sort&compress]{natbib}
\usepackage{authblk}
\usepackage{microtype}
\usepackage[hidelinks]{hyperref}
\usepackage{setspace}
\newcommand{\indep}{\perp\!\!\!\perp}
\newcommand{\1}{\mathbf{1}} 

\theoremstyle{plain}
\newtheorem{theorem}{Theorem}
\newtheorem{lemma}{Lemma}
\newtheorem{corollary}{Corollary}
\newtheorem{proposition}{Proposition}
\theoremstyle{definition}
\newtheorem{definition}{Definition}

\newif\ifshowchecks
\showchecksfalse

\title{\textbf{The Dark Matter of Clinical Data}\\[4pt]
\large An identification limit on disentangling physiology from care process in electronic health records\\[6pt]
\normalsize\textit{Cross-site invariance can falsify, but cannot identify, biological mechanism in observational clinical data}}

\author[1,2]{Mohammad Shahin\thanks{Corresponding author: \href{mailto:shahinbu@bu.edu}{shahinbu@bu.edu}}}
\author[3,4]{Krittaphas Chaisutyakorn}
\author[5]{Dimitrios Proios}
\author[6]{David Restrepo}
\author[7]{Vishala Mishra}
\author[8,5]{Boya Zhang}
\author[9]{Aadarsh Sivaraman}
\author[10,11]{Sparsh Roy}
\author[12]{Benet Fit\'{e} Abril}
\author[13]{Tina Yen-Ting Chen}
\author[14]{Hyunjung Gloria Kwak}
\author[15]{Andre Kurepa Waschka}
\author[1,16]{Vinayak Mathur}
\author[1,17]{Leo Anthony Celi}
\affil[1]{Laboratory for Computational Physiology, Massachusetts Institute of Technology, Cambridge, MA 02139, USA}
\affil[2]{Boston University School of Public Health, Boston, MA, USA}
\affil[3]{Department of Biostatistics, Harvard T.H. Chan School of Public Health, Boston, MA, USA}
\affil[4]{Siriraj Informatics and Data Innovation Center, Mahidol University, Bangkok, Thailand}
\affil[5]{Department of Radiology and Medical Informatics, Faculty of Medicine, University of Geneva, Geneva, Switzerland}
\affil[6]{Mathematics and Computer Science Laboratory (MICS), CentraleSup\'{e}lec, Universit\'{e} Paris-Saclay, Gif-sur-Yvette, France}
\affil[7]{Duke University School of Medicine, Durham, NC, USA}
\affil[8]{Institute for Medical Engineering \& Science, Massachusetts Institute of Technology, Cambridge, MA, USA}
\affil[9]{Illinois Mathematics and Science Academy, Aurora, IL, USA}
\affil[10]{Massachusetts Institute of Technology, Cambridge, MA, USA}
\affil[11]{Hopewell Valley Central High School, Pennington, NJ, USA}
\affil[12]{Department of Biomedical Informatics, Harvard Medical School, Boston, MA, USA}
\affil[13]{Reiheal Commons}
\affil[14]{Emory University, Atlanta, GA, USA}
\affil[15]{Department of Mathematics and Statistics, Elon University, Elon, NC, USA}
\affil[16]{JSPH International Boston Institute for Global Public Health, Inc., Boston, MA, USA}
\affil[17]{Division of Pulmonary, Critical Care and Sleep Medicine, Beth Israel Deaconess Medical Center, Boston, MA 02215, USA}

\date{}

\begin{document}
\maketitle
\thispagestyle{empty}

\clearpage
\begin{abstract}
\noindent\textbf{Background.} Representations learned from electronic health record (EHR) data encode patient physiology together with the clinical process that decides what is measured, when, and for whom. Whether analysis of observational data can undo this entanglement has not been formally characterized.

\noindent\textbf{Methods.} We formalize the clinical data-generating process with latent physiology, latent ordering intent, laboratory values that exist only when measured, a learned representation, and clinical environments. We analyze two strategies for separating physiology from care process, dividing the information carried by laboratory ``value'' and ``ordering'' channels within one dataset, including by partial information decomposition (PID), and cross-site tests of representation-conditional outcome invariance, and we examine what changes when a test is performed for every patient.

\noindent\textbf{Results.} When ordering depends on the analyte, its distribution among untested patients is not recoverable, so no division of value and ordering information is identified; and because the recorded value is undefined rather than missing when no test is performed, a sentinel code makes the encoded value carry all of the entropy of the ordering indicator and fixes PID for every redundancy measure that satisfies the Williams--Beer axioms and is invariant under injective re-encoding, a class that includes the standard measures. Cross-site invariance is a falsifiable implication of biological invariance taken jointly with explicit exclusion and representation-sufficiency conditions, but it is not an identification strategy. An explicit construction shows that any model in which biological invariance holds but the invariance test fails is observationally equivalent to a model in which biological invariance is false, and that passing the test does not establish it either. For binary physiology the construction applies exactly when case mix differs across sites, the representation is lossy, and physiology affects the outcome; for finite state spaces it applies for almost all parameter values. A worked example yields a 9.6-percentage-point cross-site outcome gap under an unchanged biological mechanism. Tests performed for every patient remove the ordering gate, and measured tests can restore identification: biological invariance is identified when a test summarizes the outcome-relevant physiology exactly with the same assay at every site, or, for binary physiology, when two tests with independent errors are available at sites with different case mix.

\noindent\textbf{Conclusions.} Within the model class studied here, cross-site comparison of learned clinical representations can falsify or corroborate a stated joint hypothesis but cannot identify biological invariance from observational data alone. Such analyses should be reported as tests of that joint hypothesis, with sensitivity analyses, routinely measured anchors where available, and, where possible, interventional variation in ordering.

\end{abstract}

\clearpage
\section{Introduction}\label{sec:introduction}

A laboratory value in an electronic health record (EHR) exists because a clinician, working under local protocol, staffing, and judgment, decided that the test was worth ordering at that moment. The recorded value therefore reflects several processes at once: the patient's physiology, the decision to measure, the timing of that decision, and the effects of treatments already given. A representation learned from such data compresses both the physiology being measured and the institutional attention that decided to measure it \cite{agniel,pivovarov}. The presence and timing of an order, independent of the result, predict three-year survival for most common laboratory tests \cite{agniel}, and laboratory measurement patterns, such as how often a test is repeated over time, capture the clinical context in which it was ordered, so that aggregating values across these patterns can bias analyses \cite{pivovarov}.

Prior work has shown that this entanglement arises from how routine clinical data are generated, so that the data encode the care process as well as the patient \cite{hripcsak,agniel,pivovarov,groenwold,finlayson,phelan}. If the entanglement is a property of the data-generating process, can a sufficiently well-designed method applied to the data resolve it? Two strategies suggest themselves. One decomposes, within a single dataset, the information carried by a laboratory's value from the information carried by the decision to order it. The other compares the same learned relationship across independent clinical environments, on the premise that biology does not depend on the hospital while practice does.

This paper shows that neither strategy does what it is usually taken to do, and says what each can establish instead; full proofs and an extended review are in the Supplementary Information (SI). Separating what a laboratory's value says from what the decision to order it says fails in a single dataset for a structural rather than statistical reason. When ordering depends on the analyte, its distribution among untested patients is not recoverable, so no division of the information that value and ordering carry about the outcome is identified; and encoding unmeasured values with a sentinel makes the recorded value carry the ordering decision itself. Partial information decomposition (PID), the method usually proposed for this separation, inherits both problems (Theorem~1, Corollary~1). Cross-site invariance of the representation-conditional outcome law is a falsifiable implication of biological invariance taken jointly with explicit auxiliary conditions (Theorem~2). Biological invariance itself is not identifiable from observational data: for any model in which it holds but the invariance test fails, there is an observationally identical model in which it is false (Theorem~3), and passing the test does not establish it either (Proposition~1). We characterize when this occurs, exactly for binary physiology and generically for finite state spaces (Propositions~2 and 3), and give a worked example that can be checked by hand. Measurements made for every patient remove the ordering gate, and we show when they and other measured tests can restore identification of biological invariance (Propositions~4 and~5).

The negative result turns an assumption that is usually left implicit, that cross-site agreement reflects shared biology, into an explicit identification boundary, and it says which kinds of evidence remain available. The EHR contains structured laboratory results and vital signs, medication administrations, diagnosis and procedure codes, imaging, clinician and nursing notes, and the metadata of all of these (time stamps, frequency, authorship). Selectively ordered laboratory tests are our canonical case, but the results apply to any feature whose presence or timing is decided by the care process; data recorded by protocol for every patient are the limiting case without a selective-measurement channel, and Section~\ref{sec:routine} shows what they make possible.

\section{Background and related work}\label{sec:background-and-related-work}

\textbf{Selective measurement in EHR data.} EHR data are shaped by clinical decision-making as well as by disease. The presence and timing of an order carry prognostic information independent of its result \cite{agniel}; measurement patterns over time capture the clinical context of a test, and pooling values across them can bias analyses \cite{pivovarov}; missingness reflects what clinicians believed and did \cite{groenwold}; the intensity of contact with the health system shapes which variables are observed at all \cite{phelan,goldstein,sauer}; and documentation behavior is itself a clinical signal, as in the CONCERN early warning system, which predicts deterioration from real-time patterns of nursing surveillance documentation and was evaluated in a multisite cluster-randomized trial \cite{rossetti}. Using a proxy shaped by unequal access as a stand-in for need can produce inequitable algorithms \cite{obermeyer}, and encoding social and institutional patterns in clinical algorithms more broadly can produce or amplify inequity \cite{vyas,gianfrancesco}. This literature supplies our empirical premise: selective measurement is pervasive and clinically meaningful, and it is not random missingness in Rubin's sense \cite{rubin}.

\textbf{Dataset shift and invariance.} Embedding methods that integrate sparse, heterogeneous EHR data can improve prediction \cite{kauffman}, but compression does not separate the origins of the compressed signal. Dataset shift in clinical artificial intelligence (AI) arises from changes in practice as well as in populations \cite{finlayson}, and a model's robustness depends on which causal mechanisms it relies on \cite{subbaswamy}. Invariant causal prediction and invariant risk minimization use stability of a relationship across environments as evidence of causal structure under stated assumptions \cite{peters,arjovsky}, anchor regression uses heterogeneity across environments to obtain predictors robust to bounded shifts \cite{rothenhausler}, and work on domain adaptation recovers a changed conditional distribution from unlabeled target-domain data only under structural restrictions on how it changes \cite{zhang}.

\textbf{Identification and information decomposition.} Transportability theory, built on the structural causal model framework \cite{pearl}, characterizes when causal relationships carry across environments \cite{pearlbareinboim,bareinboim}, and partial identification provides bounds when point identification fails \cite{manski90,manski03}. PID splits the information that two sources carry about a target into redundant, unique, and synergistic parts, defined through axioms rather than a single formula \cite{wb,bertschinger,griffith}. Recent work corrects PID estimates for missing modalities by inverse-probability weighting under missingness at random \cite{choi}. That correction targets values that exist but were not observed; Theorem~1 concerns laboratory values that do not exist, for which there is no complete-data distribution to recover.

\textbf{Relation to known identification results.} The non-identifiability we establish has close relatives. Read \(E\) as an exposure, \(Z^{\star}\) as a mediator, and \(Z\) as a mismeasured version of it: measurement error in a mediator biases estimates of direct effects \cite{vanderweele2012}, and an association that remains after adjusting for a mismeasured mediator need not reflect a direct effect \cite{lecessie}. The cross-site gap of Proposition~2 is an instance of this. Effects can be restored from proxies of an unmeasured confounder when the measurement process is known or constrained \cite{kurokipearl}, or when at least two proxies satisfy rank or completeness conditions \cite{miao}, and latent-class parameters are generically identifiable from several conditionally independent indicators \cite{allman}, as in the estimation of diagnostic error rates without a gold standard \cite{huiwalter}. The model class studied here imposes none of these restrictions. Structural undefinedness (Section~\ref{sec:structural-undefinedness}) parallels outcomes truncated by death, which are undefined rather than missing \cite{zhangrubin}. Our contribution is therefore not the existence of non-identifiability as such, but (i) a separation of what cross-site comparison of learned EHR representations can falsify from what it can identify, with the auxiliary assumptions named; (ii) exact and generic characterizations of when the construction applies; (iii) a calculable sensitivity model; and (iv) reporting recommendations for EHR research.

\textbf{The gap.} Even within one health system, similar observed EHR states can arise from different underlying physiology. Across systems, differences in measurement, test ordering, documentation, and practice add further variation to what is observed, so an observed cross-site difference need not reflect a difference in biology; it may arise from who the patients are or from how they are observed and treated. Prior work has documented this ambiguity empirically without a formal identification result \cite{agniel,pivovarov,phelan}, or has invoked general non-identifiability intuitions from causal inference and domain adaptation without a construction specific to EHR measurement \cite{peters,zhang}. To our knowledge, no clinical informatics work has explicitly constructed distinct data-generating processes, one with and one without cross-site biological invariance, that produce the same observed EHR data under selective measurement. We provide that construction, extend it to learned representations and the full observed record, and characterize when it applies.

\section{Formal framework}\label{sec:formal-framework}

\subsection{Variables and data-generating process}\label{sec:variables-and-data-generating-process}

We write \(Z^{\star}\) for the latent, complete physiological state; \(Z\) for a learned representation, a fixed measurable function of the observed record (for example, a frozen encoder); \(E\) for the clinical environment (hospital or health system), taking values in a finite set \(\mathcal{E}\) with \(p(e)>0\); \(S^{\star}\) for the latent ordering intent; \(C\) for censoring before measurement (death, discharge, comfort care); \(S = S^{\star}(1-C)\) for the indicator that the test was performed; \(V\) for the recorded laboratory value, defined only on \(\{S=1\}\); and \(Y\) for the outcome. Censoring and outcome are distinct: \(C\) refers to the measurement window, whereas \(Y\) is assessed over a later horizon. \(X \indep W \mid U\) denotes conditional independence. Invariance statements are made on the shared support \(\mathcal{Z}_{\mathrm{sh}} := \{z : p(z \mid e) > 0 \text{ for all } e\}\).

\begin{figure}[t]
\centering
\includegraphics[width=0.78\linewidth]{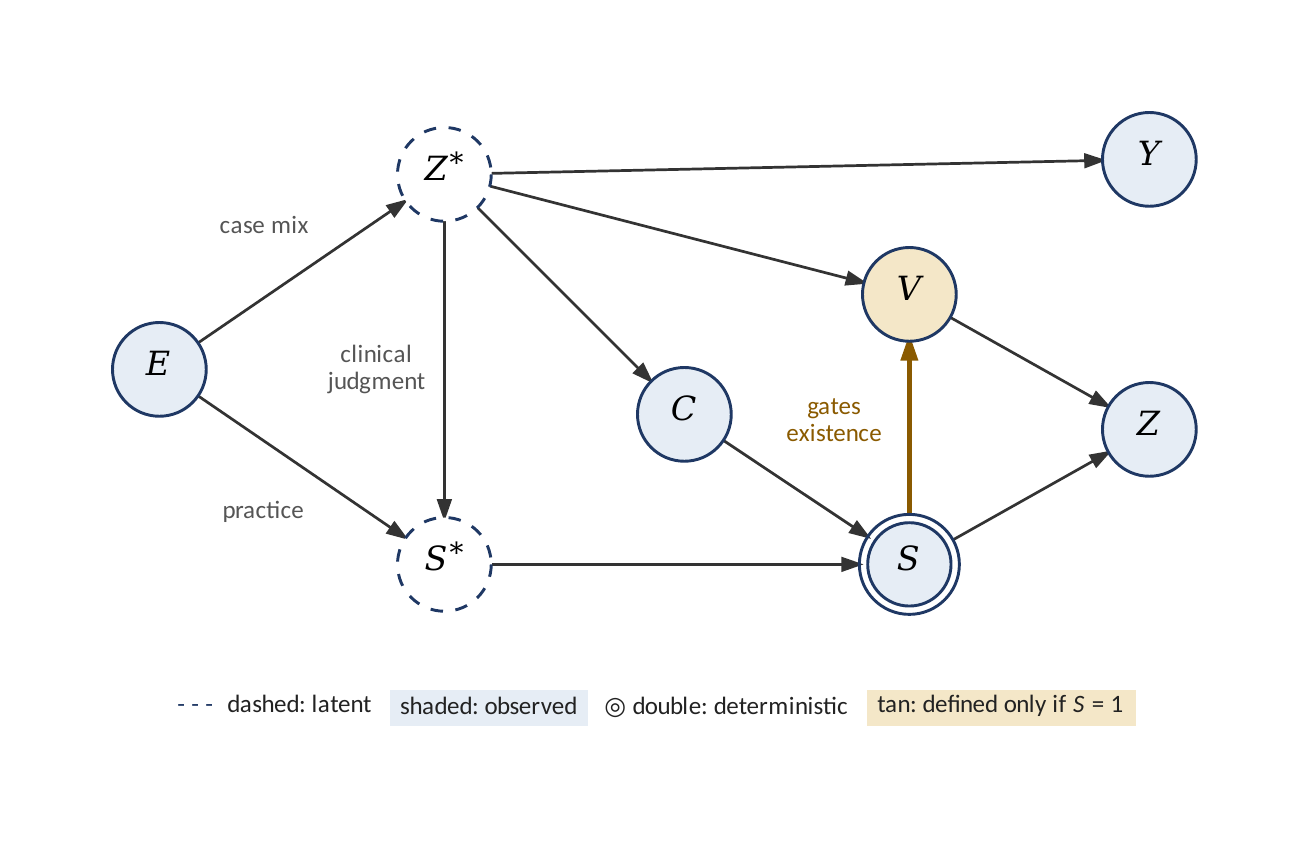}
\caption{\textbf{Clinical data-generating process.} Dashed circles are latent; shaded circles are observed; the double circle marks a deterministic node ($S = S^{\star}(1-C)$); the tan node $V$ exists only when $S=1$. The environment enters the record twice, through case mix ($E \to Z^{\star}$) and through practice ($E \to S^{\star}$), so any representation $Z$ computed from the record inherits both physiology and care process. The absent edge $E \to Y$ encodes Assumption~4, and the absent edges $S \to Y$ and $Z \to Y$ encode Assumption~6a (both stated in Section~\ref{sec:assumptions}).}
\label{fig:dgp}
\end{figure}

Figure~\ref{fig:dgp} shows the data-generating process. The environment affects both who the patients are (case mix, \(E \to Z^{\star}\)) and how they are managed (practice, \(E \to S^{\star}\)). The measurement indicator gates the existence of the recorded value, and the representation is computed from the record. The graph omits the edge \(E \to Y\) because that absence is the hypothesis under investigation.

\subsection{Structural undefinedness}\label{sec:structural-undefinedness}

\begin{figure}[t]
\centering
\includegraphics[width=\linewidth]{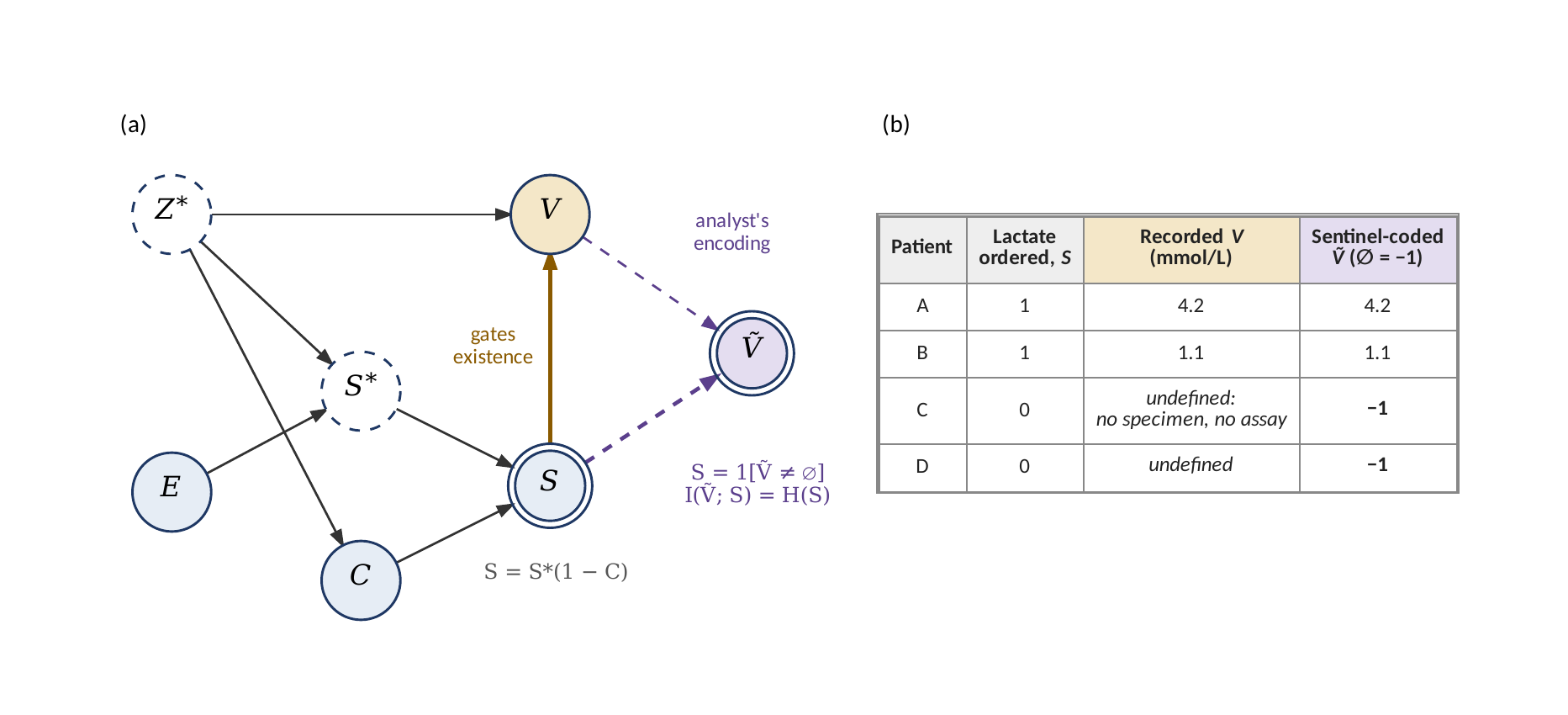}
\caption{\textbf{Selective measurement and structural undefinedness.} (a) Ordering intent responds to physiology and site practice (the case-mix edge $E \to Z^{\star}$ of Figure~\ref{fig:dgp} is omitted); $S$ gates whether $V$ exists. The analyst's encoding (purple, dashed) maps $(S,V)$ to $\tilde V$, from which $S$ can be read off exactly. (b) For patients C and D no specimen was drawn; their value is undefined rather than unknown, and the sentinel $-1$ records only that no lactate was ordered.}
\label{fig:selective}
\end{figure}

The physiological quantity that a test would assay, such as circulating lactate, is part of \(Z^{\star}\) and exists whether or not a test is ordered. We write \(V^{\circ}\) for the value the assay would return, so that \(V = V^{\circ}\) on \(\{S=1\}\). The recorded value \(V\) is different: it is the output of a measurement event, with a specimen, an assay, and a time stamp. When no test is performed there is no such event, so \(V\) has no value at all: it is \emph{structurally undefined}. This distinguishes it from a latent quantity (which exists but is never recorded) and from a missing value (which was generated but lost), to which standard missing-data theory applies \cite{rubin} (SI Table S2). Analysts usually make \(V\) defined everywhere with a sentinel: for a symbol \(\varnothing\) outside the range of \(V\) (for lactate, \(-1\) serves),
\begin{equation}\label{eq:1}
\tilde V := V \text{ on } \{S=1\}, \qquad \tilde V := \varnothing \text{ on } \{S=0\}.
\end{equation}
Because \(\tilde V = \varnothing\) exactly when the test was not performed, the encoded value contains the ordering decision (Figure~\ref{fig:selective}).

Eq.~\eqref{eq:1} is the proxy variable of the missingness graphs (m-graphs) of Mohan and Pearl \cite{mohan2013,mohanpearl2021}. In their notation \(V^{\circ}\) is a partially observed variable, \(R_{V^{\circ}} = 1 - S\) is its missingness mechanism (\(R = 0\) when the value is revealed), and \(\tilde V\) is its proxy, equal to \(V^{\circ}\) when \(R_{V^{\circ}} = 0\) and to the symbol \(m\), our \(\varnothing\), otherwise. Figure~\ref{fig:selective}a can be read as an m-graph, in which the path from \(V^{\circ}\) (a component of \(Z^{\star}\)) through ordering intent to \(S\) makes the analyte an ancestor of its own missingness indicator. We keep the symbols \(S\) and \(\tilde V\) because the ordering decision is a variable of interest in its own right, not only an indicator of missingness. The two framings differ in what the empty entry denotes. The m-graph reading treats it as a value that exists but is concealed, which is the right reading for the analyte level \(V^{\circ}\), and the m-graph results on recoverability apply to it (Section~\ref{sec:single-dataset-decomposition-is-ill-posed}). We reserve ``structurally undefined'' for the recorded value \(V\), the output of a measurement event that did not occur.

\subsection{Assumptions}\label{sec:assumptions}

Table~\ref{tab:assumptions} lists the assumptions; each result states which it uses. Assumption~4 (biological invariance) is the hypothesis whose identifiability is at issue. Assumption~6 is stated in two parts because its components play distinct roles: exclusion (6a) removes a direct effect of ordering and of the representation on the outcome, and representation sufficiency (6b) requires that, within each representation cell, the distribution of true physiology is the same at every site. Exclusion has a practical consequence for mortality outcomes: a death before the specimen could be drawn would force both \(S=0\) and \(Y=1\), so ordering would carry outcome information beyond physiology and violate Assumption~6a. We therefore take \(Y\) to be measured from the end of the measurement window (a landmark design), so that early deaths are handled as a separate competing event. We do not assume \(Y \indep E \mid Z\) directly, because that is the observable implication to be tested. The graph does not imply Assumption~6b: \(Z\) is a child of \(Z^{\star}\) and does not separate \(E\) from \(Z^{\star}\). Assumptions~5 and 7 describe the usual situation in multisite cohorts: case mix differs between sites, and a finite-dimensional embedding loses information about the state it summarizes.

\begin{table}[t]
\centering
\caption{Assumptions, their role, and whether observational data can check them. Assumption~1 (censoring ignorability) is used by no result and is discussed in SI~S1.}
\label{tab:assumptions}
\small
\begin{tabularx}{\linewidth}{@{}>{\raggedright\arraybackslash}p{2.9cm}>{\raggedright\arraybackslash}p{4.1cm}>{\raggedright\arraybackslash}X>{\raggedright\arraybackslash}p{3.1cm}@{}}
\toprule
Assumption & Statement & Role & Checkable from observational data? \\
\midrule
A2 Positivity & $0<P(S=1 \mid Z^{\star}=z^{\star}, E=e)<1$ for all $(z^{\star},e)$ with $p(z^{\star},e)>0$ & Ensures $H(S)>0$; restricts completions in Theorem~1(i) & Partly: Theorem~1(ii)--(iii) need only $0<P(S=1)<1$, which is checkable; part~(i) uses the per-site condition \\
A3 Domain restriction & $V$ undefined on $\{S=0\}$ & Forces an encoding choice (Theorem~1) & A matter of definition \\
A4 Biological invariance & $p(y \mid z^{\star},e)=p(y \mid z^{\star})$ for all $e$ with $p(z^{\star} \mid e)>0$ & Hypothesis under investigation & No (Theorem~3, Proposition~1) \\
A5 Case-mix heterogeneity & $p(z^{\star} \mid e) \neq p(z^{\star} \mid e')$ for some $e \neq e'$ & Makes posteriors site-dependent & Implied when $p(z \mid e)$ varies and the channel is site-invariant \\
A6a Exclusion & $Y \indep (Z,S) \mid Z^{\star},E$ & Auxiliary condition (Theorem~2) & No \\
A6b Representation sufficiency & $Z^{\star} \indep E \mid Z$ on $\mathcal{Z}_{\mathrm{sh}}$ & Auxiliary condition (Theorem~2); generically fails under A5 and A7 & No \\
A7 Lossy representation & $Z^{\star}$ is not almost surely a function of $Z$ & Makes posteriors non-degenerate & No; holds for finite-dimensional embeddings of a richer state \\
\bottomrule
\end{tabularx}
\end{table}
\subsection{Falsification versus identification}\label{sec:falsification-versus-identification}

\begin{definition}[model class]
\(\mathcal{M}\) consists of structural causal models \(M\) on \((E, Z^{\star}, Z, Y)\), each with a latent state space \(\mathcal{Z}^{\star}_M\) and mechanisms
\[
E := f_E(U_E), \qquad Z^{\star} := f_{Z^{\star}}(E, U_{Z^{\star}}), \qquad Z := f_Z(Z^{\star}, U_Z), \qquad Y := f_Y(Z^{\star}, E, U_Y),
\]
with mutually independent exogenous variables \(U\). The mechanisms induce the Markov kernels \(p_M(e)\), \(p_M(z^{\star} \mid e)\) (case mix), \(p_M(z \mid z^{\star})\) (representation channel), and \(p_M(y \mid z^{\star}, e)\) (outcome kernel), and through them the observational distribution
\begin{equation}\label{eq:2}
p_M(y,z,e) = \sum_{z^{\star} \in \mathcal{Z}^{\star}_M} p_M(y \mid z^{\star}, e)\, p_M(z \mid z^{\star})\, p_M(z^{\star} \mid e)\, p_M(e).
\end{equation}
The graph of every \(M\) is contained in \(E \to Z^{\star} \to Z\), \(Z^{\star} \to Y\), \(E \to Y\). The latent state space is not fixed in advance, because \(Z^{\star}\) is never observed. The mechanism \(f_Z\) summarizes the path from physiology through the observed record to the representation. The class builds in Assumption~6a, because \(f_Y\) does not take \(Z\) as an argument, and holds the channel fixed across sites, which favors identification; site-dependent channels can only enlarge the set of equivalent models. Every result below depends on \(M\) only through these kernels, and, for finite or standard Borel state spaces, every family of kernels is induced by some choice of mechanisms, so we specify models by their kernels. Following common usage we call the outcome kernel the outcome mechanism.
\end{definition}

\begin{definition}[observational equivalence]
\(M_1 \sim M_2\) if \(p_{M_1}(y,z,e) = p_{M_2}(y,z,e)\) for all \((y,z,e)\).
\end{definition}

\begin{definition}[identifiability]
A property \(\mathsf{A}\) is identifiable over \(\mathcal{M}\) if \(M_1 \sim M_2\) implies that \(\mathsf{A}\) holds in \(M_1\) exactly when it holds in \(M_2\). A single equivalent pair that disagrees about \(\mathsf{A}\) refutes identifiability.
\end{definition}

Definition~3 is Pearl's general definition of identifiability \cite{pearl,pearl2009overview}, under which a quantity is identifiable given a set of assumptions if every two models that satisfy the assumptions and induce the same observational distribution agree on it, with the quantity taken to be the indicator of whether \(\mathsf{A}\) holds. All models in \(\mathcal{M}\) have graphs contained in the graph of Definition~1, in which Assumption~4 states that the edge \(E \to Y\) carries no dependence, so identifiability asks whether matching observational distributions over this common graph force agreement about that edge. Because \(Z^{\star}\) is latent, its state space is part of the model; in the worked example of Section~\ref{sec:a-worked-instance}, \(Z^{\star}\) is binary in both models of the equivalent pair, so fixing the state space does not restore identification.

\begin{definition}[testable implication]
A restriction \(R\) on the observed distribution is a testable implication of hypothesis \(\mathsf{H}\) if every model satisfying \(\mathsf{H}\) induces a distribution satisfying \(R\).
\end{definition}

Falsifiability does not imply identifiability. A hypothesis can imply an observable restriction, so that data can refute it, while an equivalent model in which it is false still exists, so that data can never confirm it. Theorem~2 and Proposition~1 show that the conjunction of Assumptions~4 and~6 is in this position; Theorem~3 and SI Remark~1 show that Assumption~4 alone is neither identifiable nor falsifiable.

\subsection{Where the claims sit in the causal hierarchy}\label{sec:layers}

Pearl's causal hierarchy distinguishes associational (Layer~1), interventional (Layer~2), and counterfactual (Layer~3) claims, and data from a lower layer generally do not determine quantities at a higher one \cite{bareinboim2022}. Every formal result in this paper is a Layer~1 statement. Assumptions~4 and~6, the restriction of Eq.~\eqref{eq:5}, the information quantities of Theorem~1, and the equivalences of Theorem~3 and Proposition~1 all concern conditional distributions of observed and latent variables. For these results the graph of Definition~1 (\(E \to Z^{\star} \to Z\), \(Z^{\star} \to Y\), \(E \to Y\)), which summarizes Figure~\ref{fig:dgp} with the ordering nodes absorbed into the representation channel, held fixed across sites as in Definition~1, and the edge \(E \to Y\) allowed, can be read as a Bayesian network, and the structural models of Definition~1 enter only through the kernels they induce. The causal reading enters at two points. Interpreting Assumption~4 as ``the same biological mechanism at every site'' is the Layer~2 claim \(p(y \mid do(z^{\star}), e) = p(y \mid do(z^{\star}))\); because every back-door path from \(Z^{\star}\) to \(Y\) begins \(Z^{\star} \leftarrow E\), in Figure~\ref{fig:dgp} and in that graph, \(p(y \mid do(z^{\star}), e) = p(y \mid z^{\star}, e)\), and the two readings coincide. The contrast in the last row of Table~\ref{tab:example}, moving a patient between sites, is an intervention on \(E\). Because non-identifiability already holds for the Layer~1 property, it holds for its causal reading as well: if \(p(y \mid z^{\star}, e)\) is not determined by the observed distribution, neither is the interventional quantity that equals it under the graph. No result requires counterfactual assumptions.

\section{Results}\label{sec:results}

\subsection{A single dataset cannot separate value from ordering}\label{sec:single-dataset-decomposition-is-ill-posed}

Dividing what a laboratory test says about the outcome into a part due to its value and a part due to the decision to order it requires a joint distribution of the outcome, the value, and the ordering indicator over the whole population. There are two candidates for the value. The analyte level \(V^{\circ}\) is defined for every patient but observed only when the test is performed. The recorded value is observed whenever it exists, but under Assumption~3 it exists only on \(\{S=1\}\) and has to be extended to the whole population by an encoding such as Eq.~\eqref{eq:1}. The next theorem shows that neither candidate supports the separation. It is stated first for the mutual informations to which any such division must add up, with no commitment to a particular decomposition method, and then for partial information decomposition (PID), the method usually proposed for this purpose.

\begin{theorem}[a single dataset cannot separate value from ordering]
Let Assumption~2 hold.

(i) \emph{Analyte level.} Let \(Y\) and \(V^{\circ}\) be discrete, suppose \(P(S=0)>0\), \(H(Y \mid S=0)>0\), and \(p(y, v^{\circ} \mid S=1) > 0\) for every \((y, v^{\circ})\), and place no restriction on how ordering depends on \(V^{\circ}\) beyond Assumptions~2 and~6a and the observed distribution of \((Y, S, \tilde V)\). That distribution determines the law of \((Y, V^{\circ})\) on \(\{S=1\}\); the assumptions require \(p(v^{\circ} \mid y, S=0)\) to be positive for every \((y, v^{\circ})\) but restrict it no further; and \(I(Y; V^{\circ} \mid S)\) takes every value in
\begin{equation}\label{eq:3a}
\bigl[\,P(S{=}1)\, I(Y;V^{\circ} \mid S{=}1),\;\; P(S{=}1)\, I(Y;V^{\circ} \mid S{=}1) + P(S{=}0)\, H(Y \mid S{=}0)\,\bigr)
\end{equation}
as \(p(v^{\circ} \mid y, S=0)\) ranges over the compatible laws, provided \(V^{\circ}\) can take at least as many values as \(Y\) (values arbitrarily close to the upper end are then attained). Hence \(I(Y; V^{\circ}, S) = I(Y;S) + I(Y; V^{\circ} \mid S)\) is not identified, and neither is at least one of the redundant, unique, and synergistic terms of any decomposition of it.

(ii) \emph{Recorded value.} If \(\tilde V\) is defined by Eq.~\eqref{eq:1}, then \(S = \1[\tilde V \neq \varnothing]\) almost surely, so \(H(S \mid \tilde V) = 0\) and
\begin{equation}\label{eq:3}
I(\tilde V; S) = H(S) > 0,
\end{equation}
whatever the joint law of \((Z^{\star}, S^{\star}, C, V)\), including when ordering is independent of physiology.

(iii) \emph{PID of the recorded value.} For every redundancy measure that satisfies the Williams--Beer axioms \cite{wb,bertschinger} and is invariant under injective re-encoding of a source, and for every outcome \(Y\),
\begin{equation}\label{eq:4}
\begin{aligned}
&\operatorname{Red}(Y; \tilde V, S) = I(Y;S), &\qquad& \operatorname{Unq}(Y; S \setminus \tilde V) = 0,\\
&\operatorname{Syn}(Y; \tilde V, S) = 0, &\qquad& \operatorname{Unq}(Y; \tilde V \setminus S) = I(Y;\tilde V) - I(Y;S).
\end{aligned}
\end{equation}
\end{theorem}

\begin{proof}[Proof sketch]
(i) The observed distribution fixes \(p(y,s)\) and \(p(y, v^{\circ} \mid S=1)\). A positive conditional law \(q(v^{\circ} \mid y)\) on \(\{S=0\}\) completes it to a joint law with full support, and the data-generating process of Figure~\ref{fig:dgp} produces each such completion while satisfying Assumptions~2 and~6a: take \(Z^{\star} = (V^{\circ}, W)\) with \(Y := W\), and let ordering intent given \(Z^{\star}\) have the positive probabilities the completion implies. Assumptions~2 and~6a force each completion to be positive wherever \(p(y, v^{\circ} \mid S=1)\) is. A positive \(q\) free of \(y\) gives \(I(Y;V^{\circ} \mid S=0) = 0\); positive \(q\) that concentrate on an injective function of \(Y\) give values approaching \(H(Y \mid S=0)\). The compatible completions form a convex set on which \(I(Y; V^{\circ} \mid S=0)\) is continuous, so every intermediate value is attained. The terms of any bivariate decomposition sum to \(I(Y;V^{\circ},S)\), so they cannot all be identified. (ii) On \(\{S=1\}\), \(\tilde V = V \neq \varnothing\); on \(\{S=0\}\), \(\tilde V = \varnothing\). Hence \(S\) is a function of \(\tilde V\), and \(H(S)>0\) by Assumption~2. (iii) Because \(S\) is a function of \(\tilde V\), the monotonicity equality condition and re-encoding invariance give \(\operatorname{Red} = I(Y;S)\); the remaining terms follow from the bivariate decomposition identities and \(I(Y;\tilde V, S) = I(Y;\tilde V)\). Full proofs in SI S3, which also shows that the measures of Williams and Beer, Bertschinger and colleagues, and Griffith and Koch \cite{wb,bertschinger,griffith} are re-encoding invariant.
\end{proof}

Part~(i) is a recoverability statement in the sense of the missingness-graph literature \cite{mohan2013,mohanpearl2021}. When the decision to order can depend on the analyte itself, directly or through the clinician's suspicion, the analyte's distribution among untested patients cannot be recovered from the data, and every quantity that depends on it inherits this: the information the value would carry about the outcome, and any way of dividing that information between value and ordering. The statement needs no axioms about how information is decomposed. It is removed only by assumptions that make the untested distribution recoverable, for example that ordering is independent of the analyte given the outcome and other observed variables, which selective ordering in the EHR is designed to violate, or by measuring every patient (Section~\ref{sec:routine}). In a small example with binary \(Y\) and \(V^{\circ}\), two completions that agree on everything observed and satisfy every assumption give \(I(Y;V^{\circ},S) = 0.247\) and \(0.556\) bits (SI S3).

Parts~(ii) and~(iii) concern the other candidate. The code for ``not measured'' is a record of the ordering decision. The axioms in part~(iii) say, informally, that redundancy does not depend on the order in which sources are listed, that a single source shares all of its information about \(Y\) with itself, and that adding a source to a collection can only reduce what the collection shares, with no reduction when the added source is a function of one already present; re-encoding invariance says that relabeling a source's values leaves the result unchanged. SI S3 states the axioms formally and checks them for the measures cited. Every admissible redundancy measure therefore attributes all of the ordering indicator's predictive content to information ``shared'' with the value, leaving ordering no unique information and no synergy, whichever measure is chosen \cite{wb,bertschinger,griffith}. If lactate is ordered mainly when a clinician suspects sepsis and unmeasured lactate is stored as \(-1\), PID will report that ``lactate'' and ``ordering'' share whatever mortality information clinician suspicion carries, even if measured lactate values add nothing. None of this makes a missingness indicator a poor predictor; it is often among the most informative features available \cite{agniel}, and handling of missingness that would be inappropriate for causal or descriptive aims can be appropriate for prediction \cite{sperrin}. The point is interpretive: a decomposition applied to a sentinel-encoded value cannot be read as separating what the value says from what the order says. Other encodings run into the same problem in a different form.

\begin{corollary}[no encoding-neutral resolution]
Under Assumptions~2 and 3, any decomposition into ``value'' and ``ordering'' contributions must (a) restrict to measured patients, on whom \(S\) is constant, so the ordering channel is removed rather than decomposed; (b) fill in \(\tilde V = W\) on \(\{S=0\}\), in which case \(I(\tilde V;S) = 0\) if and only if the law of \(W\) given \(S=0\) equals the law of \(V\) given \(S=1\), and otherwise the information about ordering is a property of the fill-in rule; or (c) posit a value for unmeasured patients, which contradicts Assumption~3 or, if the analyte level \(V^{\circ}\) is used instead, makes the decomposition depend on a distribution that is not identified (Theorem~1(i)). (Proof in SI S3.)
\end{corollary}

\subsection{Cross-site invariance is a falsifiable implication}\label{sec:cross-site-invariance-is-a-falsifiable-implication}

\begin{theorem}[testable implication of biological invariance]
Under Assumptions~4 and 6,
\begin{equation}\label{eq:5}
p(y \mid z, E=e) = p(y \mid z) \quad \text{for every } e \in \mathcal{E} \text{ and } z \in \mathcal{Z}_{\mathrm{sh}}.
\end{equation}
\end{theorem}

\begin{proof}
Fix \(e\) and \(z \in \mathcal{Z}_{\mathrm{sh}}\). Then
\begin{align*}
p(y \mid z, e) &= \sum_{z^{\star}} p(y \mid z^{\star}, z, e)\, p(z^{\star} \mid z, e)
\overset{\text{(6a)}}{=} \sum_{z^{\star}} p(y \mid z^{\star}, e)\, p(z^{\star} \mid z, e)\\
&\overset{\text{(A4)}}{=} \sum_{z^{\star}} p(y \mid z^{\star})\, p(z^{\star} \mid z, e)
\overset{\text{(6b)}}{=} \sum_{z^{\star}} p(y \mid z^{\star})\, p(z^{\star} \mid z),
\end{align*}
which does not depend on \(e\).
\end{proof}

Each assumption removes the site dependence of exactly one factor. Theorem~2 converts a statement about latent physiology into a restriction on observables that can be tested in any cohort with two or more sites, for example by testing \(Y \indep E \mid Z\) within strata of \(Z\). By contraposition, a reliable violation shows that at least one of Assumptions~4, 6a, and 6b fails. The implication runs in one direction only; passing the test establishes none of the three.

\begin{proposition}[passing the test does not establish biological invariance]
Let \(Y\) be binary and let \(P_{\mathrm{obs}}\) be any distribution of \((Y,Z,E)\) with finite \(\mathcal{Z}\) that satisfies Eq.~\eqref{eq:5}, with \(P(Y=1 \mid z, e) \in (0,1)\) wherever \(P(z \mid e)>0\) and \(\mathcal{Z}_{\mathrm{sh}}\) non-empty. Then some \(M_3 \in \mathcal{M}\) generates \(P_{\mathrm{obs}}\), satisfies Assumptions~6a, 6b, and 7, and violates Assumption~4.
\end{proposition}

\begin{proof}[Proof idea]
Augment the latent state to \((Z,U)\) with an independent binary \(U\), and add site-specific perturbations \(\pm\delta_e\) to the outcome mechanism that average to zero over \(U\) within every representation cell (SI S4).
\end{proof}

Assumption~4 on its own has no testable implication over \(\mathcal{M}\): any distribution of \((Y,Z,E)\) on finite alphabets is generated by a site-invariant mechanism if \(Z^{\star}\) is taken to be \((Z,Y)\) (SI Remark~1). A passed test therefore constrains a margin of the joint distribution and says nothing on its own about the mechanism beneath it.

\subsection{Non-identifiability of biological invariance}\label{sec:non-identifiability-of-biological-invariance}

The central construction replaces the latent physiology of a model by the representation itself and moves all remaining site dependence into the outcome mechanism.

\begin{lemma}[collapsing construction]
For \(M_1 \in \mathcal{M}\), define \(M_2\) by \(\mathcal{Z}^{\star}_{M_2} := \mathcal{Z}\), \(p_{M_2}(e) := p_{M_1}(e)\), and
\begin{equation}\label{eq:6}
\begin{aligned}
p_{M_2}(z^{\star} \mid e) &:= p_{M_1}(Z = z^{\star} \mid e), \qquad p_{M_2}(z \mid z^{\star}) := \1[z = z^{\star}],\\
p_{M_2}(y \mid z^{\star}, e) &:= p_{M_1}(y \mid Z = z^{\star}, e).
\end{aligned}
\end{equation}
Where \(p_{M_1}(Z = z^{\star} \mid e) = 0\), the last conditional is undefined; set \(p_{M_2}(y \mid z^{\star}, e)\) to its value at any site where \(z^{\star}\) occurs (or to any fixed distribution if \(z^{\star}\) occurs at no site), a choice that does not affect \(p_{M_2}(y,z,e)\). Then \(M_2 \in \mathcal{M}\) and \(M_2 \sim M_1\).
\end{lemma}

\begin{proof}
Substituting Eq.~\eqref{eq:6} into Eq.~\eqref{eq:2}, the identity channel collapses the sum to \(z^{\star} = z\), giving \(p_{M_2}(y,z,e) = p_{M_1}(y \mid z, e)\, p_{M_1}(z \mid e)\, p(e) = p_{M_1}(y,z,e)\).
\end{proof}

\begin{theorem}[non-identifiability of biological invariance]
Let \(M_1 \in \mathcal{M}\) satisfy Assumption~4 and let \(M_2\) be given by Lemma~1. Then \(M_2 \sim M_1\); \(M_2\) satisfies Assumptions~6a and 6b; and \(M_2\) satisfies Assumption~4 if and only if, for every \(z\), \(p_{M_1}(y \mid z, e)\) is the same at all sites \(e\) with \(p_{M_1}(z \mid e) > 0\) (with two sites, if and only if it does not depend on \(e\) on \(\mathcal{Z}_{\mathrm{sh}}\)). Consequently, whenever \(M_1\) satisfies Assumption~4 but its representation-conditional outcome law varies across sites, \(M_1\) and \(M_2\) are observationally equivalent and disagree about Assumption~4, which is therefore not identifiable over \(\mathcal{M}\). When the variation occurs on \(\mathcal{Z}_{\mathrm{sh}}\), \(M_1\) violates Assumption~6b while \(M_2\) satisfies it, so a failed invariance test cannot be attributed to a failure of Assumption~4 rather than of Assumption~6.
\end{theorem}

\begin{proof}
In \(M_2\), \(Z = Z^{\star}\) almost surely, so 6a and 6b hold trivially, and by Eq.~\eqref{eq:6} the mechanism is free of \(e\) exactly under the stated condition. If \(M_1\) satisfied 6b, Theorem~2 would make \(p_{M_1}(y \mid z,e)\) invariant, a contradiction.
\end{proof}

Non-identifiability therefore reduces to one question: do models exist that satisfy Assumption~4 yet have site-varying representation-conditional outcome laws? Under Assumption~4 and Bayes' rule,
\begin{equation}\label{eq:7}
p_{M_1}(y \mid z, e) = \sum_{z^{\star}} p_{M_1}(y \mid z^{\star})\, p_{M_1}(z^{\star} \mid z, e), \qquad p_{M_1}(z^{\star} \mid z, e) \propto p_{M_1}(z \mid z^{\star})\, p_{M_1}(z^{\star} \mid e).
\end{equation}
The mechanism is shared; all site dependence enters through the posterior, and there only through the case-mix prior. Conditioning on \(Z\) does not block the path \(E \to Z^{\star} \to Y\), because \(Z\) is a lossy descendant of \(Z^{\star}\) rather than a node on the path (SI Figure S1); only conditioning on \(Z^{\star}\) itself would block it, and \(Z^{\star}\) is unobserved.

\begin{proposition}[exact conditions for binary physiology]
Let \(M_1\) satisfy Assumption~4 with \(Z^{\star} \in \{0,1\}\), and write \(\mu_k := P(Y=1 \mid Z^{\star}=k)\), \(q_e := P(Z^{\star}=1 \mid e)\), \(a_z := P(Z=z \mid Z^{\star}=1)\), and \(b_z := P(Z=z \mid Z^{\star}=0)\). Then, for \(z\) with \(p_{M_1}(z \mid e) > 0\) and \(p_{M_1}(z \mid e') > 0\),
\begin{equation}\label{eq:8}
\begin{aligned}
P(Y{=}1 \mid z, e) &= \mu_0 + (\mu_1 - \mu_0)\, \pi_e(z), \qquad \pi_e(z) := \frac{a_z q_e}{a_z q_e + b_z (1-q_e)},\\
\Delta_{e,e'}(z) &= (\mu_1 - \mu_0)\,[\pi_e(z) - \pi_{e'}(z)],
\end{aligned}
\end{equation}
and the cross-site gap \(\Delta_{e,e'}(z)\) is non-zero if and only if (a) \(\mu_1 \neq \mu_0\), physiology affects the outcome; (b) \(q_e \neq q_{e'}\), case mix differs; and (c) \(a_z > 0\) and \(b_z > 0\), the representation is lossy at \(z\).
\end{proposition}

\begin{proof}
Eq.~\eqref{eq:8} is Eq.~\eqref{eq:7} with binary \(Z^{\star}\). If \(a_z b_z = 0\), \(\pi_e(z) \in \{0,1\}\) for every \(e\). Otherwise \(d\pi/dq = a_z b_z/[a_z q + b_z(1-q)]^2 > 0\), so \(\pi_e(z) = \pi_{e'}(z)\) if and only if \(q_e = q_{e'}\).
\end{proof}

\begin{proposition}[genericity for finite state spaces]
With finite alphabets \(\mathcal{Z}^{\star}, \mathcal{Z}, \mathcal{Y}, \mathcal{E}\), each with at least two elements, and \(p(e)>0\) fixed, the set of parameters of models satisfying Assumption~4 for which \(p(y \mid z, e)\) is invariant in \(e\) on \(\mathcal{Z}_{\mathrm{sh}}\) has Lebesgue measure zero. (Proof in SI S4: invariance requires a polynomial in the parameters to vanish, and the example of Section~\ref{sec:a-worked-instance} shows that the polynomial is not identically zero.)
\end{proposition}

Exact cancellation of case-mix differences within every representation cell is therefore non-generic, in the same sense as violations of faithfulness \cite{meek,sgs}, so the models Theorem~3 requires are typical rather than contrived. The same reasoning applies in reverse to Proposition~1, whose model \(M_3\) relies on exact cancellation of site-specific perturbations within every representation cell, so that a genericity argument favors the conjunction of Assumptions~4 and~6b when the test is passed. Genericity bears on which explanation is typical, not on what the data identify. Measure zero concerns exact cancellation; approximate cancellation, which produces a gap too small to detect at a given sample size, is not excluded, and the sensitivity analysis of Section~\ref{sec:a-worked-instance} is the tool for that case. Two extensions close the obvious escape routes (SI S4). The equivalent model can be chosen to satisfy Assumptions~5 and 7 as well, by augmenting its latent state with an independent coordinate, so restricting attention to lossy, heterogeneous models does not restore identifiability. And applying Lemma~1 to the full observed record \(W := (Z, \tilde V)\) yields a model equivalent on \((y,z,s,v,e)\) (SI Corollary~2). Ordering indicators and values can sharpen the posterior over physiology. Without further restrictions on how the record depends on physiology, they remove the ambiguity only if they pin down its outcome-relevant component; restrictions such as conditionally independent proxies or a bounded latent state space can bring latent-variable identification results into play, but they are assumptions in their own right. When ordering practice differs by site, ordering indicators and values add a further source of site dependence.

\begin{figure}[t]
\centering
\includegraphics[width=0.92\linewidth]{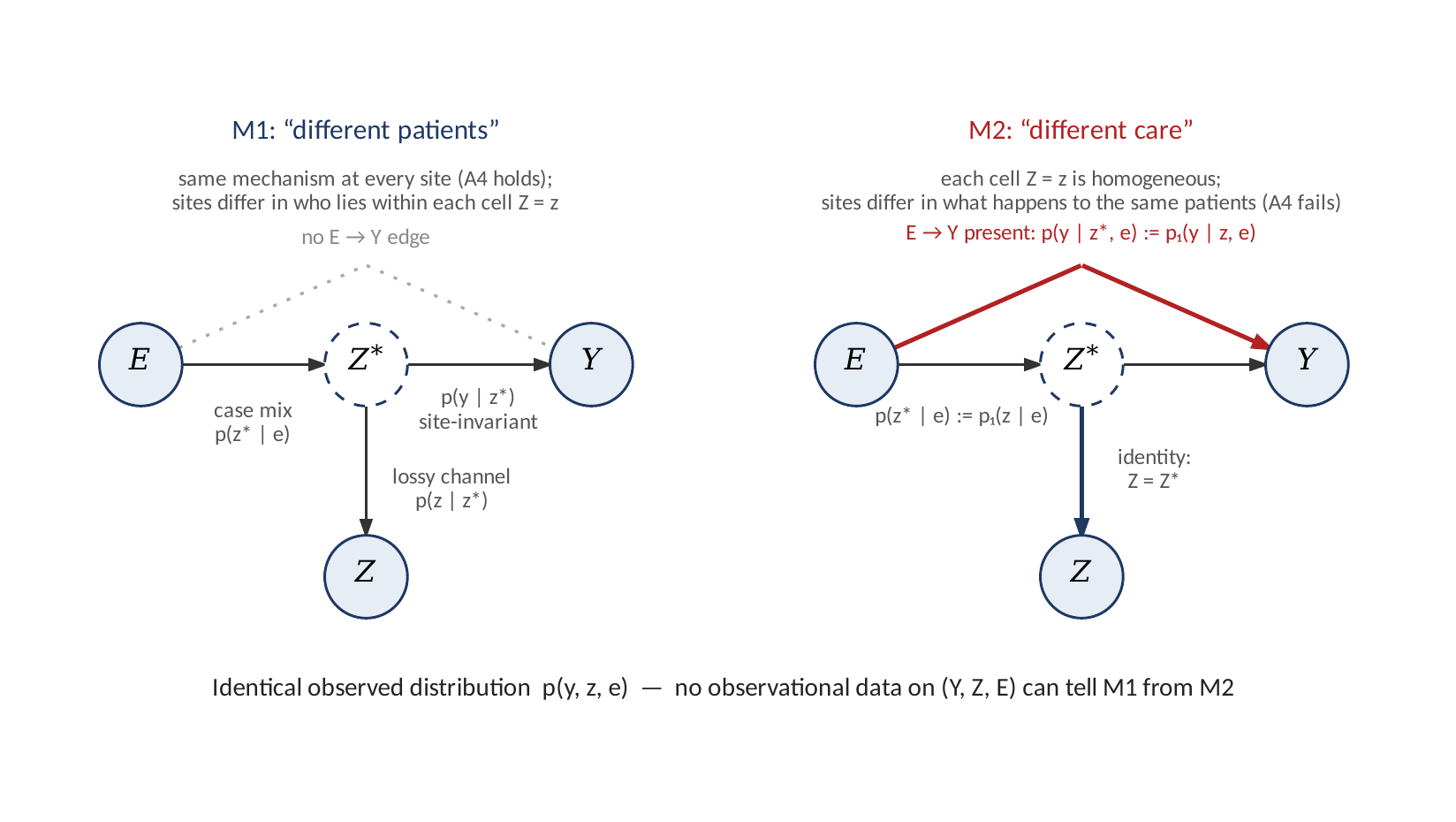}
\caption{\textbf{Observational equivalence.} Left, $M_1$: biological invariance holds (no $E \to Y$ edge), case mix differs by site, and the representation is lossy, so the posterior over physiology within a cell $Z=z$ depends on the site. Right, $M_2$ (Lemma~1): the latent state is the representation itself and the outcome mechanism is site-specific ($E \to Y$ present). Both induce the same $p(y,z,e)$.}
\label{fig:equivalence}
\end{figure}

Figure~\ref{fig:equivalence} contrasts the two worlds: in \(M_1\) the sites have different patients within each representation cell; in \(M_2\) they give the same patients different care. The joint distribution of \((Y,Z,E)\) cannot distinguish them.

\subsection{A worked instance}\label{sec:a-worked-instance}

Let \(E, Z^{\star}, Z, Y \in \{0,1\}\). For concreteness, read \(E\) as two hospital systems, \(Z^{\star}\) as true sepsis status, \(Z\) as a high-risk flag from a lactate-based embedding, and \(Y\) as 30-day mortality (the numbers are stylized and chosen for arithmetic transparency; a conditional mortality of 0.8 is higher than is typical in sepsis cohorts). In \(M_1\) (Table~\ref{tab:example}), case mix differs (\(q_0 = 0.3\), \(q_1 = 0.7\)), the channel is identical at both sites with 10\% error (\(a_1 = 0.9\), \(b_1 = 0.1\)), and the mechanism is site-invariant (\(\mu_0 = 0.2\), \(\mu_1 = 0.8\)). Then \(P(Z=1 \mid e) = 0.34\) and \(0.66\), and among patients with \(Z=1\) the posterior probability of the high-risk state is \(27/34 \approx 0.794\) at site 0 and \(21/22 \approx 0.955\) at site 1. By Eq.~\eqref{eq:8}, the observed mortality in the \(Z=1\) cell is \(23/34 \approx 0.676\) and \(17/22 \approx 0.773\): a gap of \(18/187 \approx 9.6\) percentage points, with an identical gap in the \(Z=0\) cell. Lemma~1 then gives \(M_2\), whose site-specific mechanism equals these observed rates. Both models induce the same joint distribution (Table~\ref{tab:example}; exact fractions in SI S5).

\begin{table}[t]
\centering
\caption{$M_1$ versus $M_2$: different structures, identical observations.}
\label{tab:example}
\small
\begin{tabularx}{\linewidth}{@{}>{\raggedright\arraybackslash}X>{\raggedright\arraybackslash}p{3.6cm}>{\raggedright\arraybackslash}p{3.6cm}@{}}
\toprule
Quantity & $M_1$ (``different patients'') & $M_2$ (``different care'') \\
\midrule
Latent state & $Z^{\star}$ distinct from $Z$ & $Z^{\star} = Z$ \\
Case mix $P(Z^{\star}=1 \mid E=0)$, $P(Z^{\star}=1 \mid E=1)$ & 0.30, 0.70 & 0.34, 0.66 \\
Channel $P(Z=1 \mid Z^{\star}=1)$, $P(Z=1 \mid Z^{\star}=0)$ & 0.9, 0.1 (lossy) & 1, 0 (identity) \\
Mechanism $P(Y=1 \mid Z^{\star}=1)$ at $E=0$, $E=1$ & 0.800, 0.800 & 0.676, 0.773 \\
Mechanism $P(Y=1 \mid Z^{\star}=0)$ at $E=0$, $E=1$ & 0.200, 0.200 & 0.227, 0.324 \\
Assumption~4 / Assumption~6b & Holds / fails & Fails / holds \\
$P(Y,Z \mid E=0)$ at $(1,1),(0,1),(1,0),(0,0)$ & 0.23, 0.11, 0.15, 0.51 & 0.23, 0.11, 0.15, 0.51 \\
$P(Y,Z \mid E=1)$ at $(1,1),(0,1),(1,0),(0,0)$ & 0.51, 0.15, 0.11, 0.23 & 0.51, 0.15, 0.11, 0.23 \\
Effect of moving a $Z^{\star}=1$ patient from site 0 to site 1 & None & $+9.6$ percentage points \\
\bottomrule
\end{tabularx}
\end{table}
The gap is detectable at ordinary sample sizes: about 340 patients per site in the \(Z=1\) stratum give 80\% power to detect it at \(\alpha = 0.05\). More observational data sharpen the estimate of the gap but cannot adjudicate between \(M_1\) and \(M_2\).

Eq.~\eqref{eq:8} nonetheless supports a calculable sensitivity analysis. If an analyst posits a symmetric, site-invariant representation error rate \(\varepsilon\), the case mix is recovered as \(q_e = [P(Z=1 \mid e) - \varepsilon]/(1-2\varepsilon)\), and because \(|\mu_1 - \mu_0| \le 1\), biological invariance can explain a gap of at most \(|\pi_1(z) - \pi_0(z)|\) in cell \(z\). In the example this maximum is 0.029 at \(\varepsilon = 0.02\), 0.075 at \(\varepsilon = 0.05\), and 0.160 at \(\varepsilon = 0.10\), so the 9.6-point gap in either cell is compatible with Assumption~4 only if \(\varepsilon \gtrsim 0.063\). This one-cell bound is a necessary condition. Using both cells is sharper. The binary model with a symmetric, site-invariant channel is a latent-class model of the kind used to estimate diagnostic error rates without a gold standard \cite{huiwalter}: under Assumption~4 it has five parameters \((q_0, q_1, \varepsilon, \mu_0, \mu_1)\) for six observed degrees of freedom, so it imposes a testable restriction and, when that restriction holds, identifies \(\varepsilon \in [0,\tfrac12)\) provided \(q_0 \neq q_1\) and \(\mu_0 \neq \mu_1\). In the example the four cell rates are reproduced exactly only at \(\varepsilon = 0.10\), which recovers \(M_1\). This does not contradict Theorem~3: \(\varepsilon\) is identified only conditionally on Assumption~4, and \(M_2\), which is binary with a symmetric channel (\(\varepsilon = 0\)) but a site-specific mechanism, fits the same data exactly. The calculation is therefore a joint test of Assumption~4 and the parametric latent model. An external estimate of the error rate, for example from a validation study of the embedding against adjudicated sepsis status, is what turns it into evidence about Assumption~4: an error rate credibly different from 0.10 would, within this model, count against it.

\subsection{Routine measurements remove the gate}\label{sec:routine}

Everything above turns on the ordering indicator. Some measurements have none in practice: a basic metabolic panel or complete blood count drawn at nearly every inpatient or intensive care admission is measured for every patient. For such a test \(S \equiv 1\), the recorded value exists for everyone, and its distribution is observed without selection. Theorem~1(i) and~(ii) no longer apply, because \(P(S=0)=0\) and \(H(S)=0\); the site-specific ordering pathway into the representation (SI Corollary~2) disappears; and every value the test can take is observed at every site. Routine measurements therefore show what the data look like without the gate. They also make it possible to state conditions under which the cross-site comparison identifies biological invariance. In Figure~\ref{fig:dgp} a measured value is a child of physiology with no edge to the outcome, so \(Y \indep V \mid (Z^{\star}, E)\); two routes to identification follow.

\begin{proposition}[an exact routine summary identifies biological invariance]
Let a test be performed for every patient at every site, with value \(V = g(Z^{\star})\) for a function \(g\) that is the same at every site, and let every physiological state occur at every site. Suppose the outcome depends on physiology only through this value: there are kernels \(h_e\) with \(p(y \mid z^{\star}, e) = h_e(y \mid g(z^{\star}))\). Then Assumption~4 holds if and only if \(p(y \mid v, e)\) does not depend on \(e\) for every value \(v\) of \(V\).
\end{proposition}

\begin{proof}
For each \(v\) and \(e\), \(p(y \mid v, e) = \sum_{z^{\star}:\, g(z^{\star}) = v} h_e(y \mid v)\, p(z^{\star} \mid v, e) = h_e(y \mid v)\). Assumption~4 states that \(h_e(y \mid g(z^{\star}))\) does not depend on \(e\) for any \(z^{\star}\); because every state occurs at every site, this is the statement that \(p(y \mid v, e) = h_e(y \mid v)\) does not depend on \(e\) for any value \(v\).
\end{proof}

\setcounter{corollary}{2}
\begin{corollary}[a routine reference standard calibrates the sensitivity analysis]
In the binary model of Proposition~2 with a site-specific mechanism \(\mu_{k,e} := P(Y=1 \mid Z^{\star}=k, E=e)\), suppose a routine test satisfies \(V = Z^{\star}\) at every site. Then the case mix \(q_e = P(V=1 \mid e)\), the representation's error rates \(a_z = P(Z=z \mid V=1)\) and \(b_z = P(Z=z \mid V=0)\), and \(\mu_{k,e} = P(Y=1 \mid V=k, e)\) are identified, so Assumption~4 can be tested directly and the cross-site gap of Eq.~\eqref{eq:8} splits into its case-mix and mechanism parts.
\end{corollary}

In the worked example, such a reference standard would show case mix of \(0.3\) and \(0.7\), error rates of \(0.1\), and a mechanism that is the same at both sites, settling the question between \(M_1\) and \(M_2\) that the distribution of \((Y, Z, E)\) leaves open. An exact summary of the outcome-relevant state is a strong requirement. The second route replaces it with two imperfect tests, and uses the case-mix difference between sites, the source of the ambiguity in Proposition~2, as the source of identification.

\begin{proposition}[two tests and a case-mix difference identify biological invariance]
Let physiology be binary, \(Z^{\star} \in \{0,1\}\), at two sites with case mix \(q_0 \neq q_1\), both in \((0,1)\). Let two binary tests \(V_1\) and \(V_2\) be performed for every patient at both sites, with sensitivities and specificities that are the same at both sites and sum to more than one for each test, and let \(V_1\), \(V_2\), and \(Y\) be conditionally independent given \((Z^{\star}, E)\). Then the case mix, both tests' error rates, and the site-specific mechanism \(\mu_{k,e} = P(Y=1 \mid Z^{\star}=k, E=e)\) are identified from the distribution of \((Y, V_1, V_2, E)\); in particular, Assumption~4 is identified.
\end{proposition}

\begin{proof}
At each site, \(p(v_1, v_2 \mid e)\) follows a latent-class model with prevalence \(q_e\) and error rates common to both sites. With two populations of different prevalence and conditionally independent tests, these six parameters are identified from the six observed degrees of freedom up to swapping the labels of \(Z^{\star}\) \cite{huiwalter}, and the condition that sensitivity and specificity sum to more than one fixes the labels. For \(v_1 \in \{0,1\}\), \(P(Y{=}1, V_1{=}v_1 \mid e) = \sum_{k} P(Z^{\star}{=}k \mid e)\, P(V_1{=}v_1 \mid Z^{\star}{=}k)\, \mu_{k,e}\) is a linear system in \((\mu_{0,e}, \mu_{1,e})\) whose determinant is \(\pm\, q_e(1-q_e)(\mathrm{sens}_1 + \mathrm{spec}_1 - 1) \neq 0\), so it has a unique solution.
\end{proof}

Neither proposition contradicts Theorem~3. Proposition~4 assumes that the measured value captures all of the outcome-relevant physiology, which excludes the lossy models on which Theorem~3 relies; Proposition~5 assumes binary physiology and conditional independence of the tests and the outcome given it, under which the latent-class structure is identifiable. The observational distribution cannot check these assumptions, and the identification comes from them. They are, however, assumptions about assays, which laboratory knowledge can support, rather than about how sites treat patients. Because exclusion (Assumption~6a) makes the outcome mechanism the same among tested and untested patients, both propositions also apply within the tested subpopulation when ordering depends only on physiology and site, as in Figure~\ref{fig:dgp}; routine measurement supplies what that application would otherwise have to assume: that every state and value is observed at every site, that, for Proposition~5, the case mix among tested patients still differs between sites (site-dependent ordering can equalize it), and that the representation carries no site-specific ordering. A numerical check of Proposition~5 is included with the code accompanying this article.

\section{Discussion}\label{sec:discussion}

\subsection{What researchers should do differently}\label{sec:what-researchers-should-do-differently}

Four changes follow. A claim that a representation ``captures biology'' because its relationship to outcomes is stable across sites should be restated as a claim about the joint hypothesis of Assumptions~4 and 6. Failed and passed tests should be reported as ``inconsistent with'' or ``consistent with'' that hypothesis rather than as evidence of biological difference or sameness. Sentinel codes and missingness indicators should be treated as carriers of care-process information. Cross-site analyses should report sensitivity analyses over interpretable parameters, such as case-mix contrasts and representation error rates, and should state whether interventional or quasi-experimental variation in ordering was available. Table~\ref{tab:reporting} turns these points into reporting items.

\begin{table*}[t]
\centering
\caption{What each analysis can establish from observational data, and what should be reported.}
\label{tab:reporting}
\small
\begin{tabularx}{\linewidth}{@{}>{\raggedright\arraybackslash}p{2.9cm}>{\raggedright\arraybackslash}X>{\raggedright\arraybackslash}X>{\raggedright\arraybackslash}X>{\raggedright\arraybackslash}p{2.3cm}@{}}
\toprule
Analysis & Can establish & Cannot establish & Report & Basis \\
\midrule
Dividing information between value and ordering in one dataset (including PID) & Nothing about separate physiological and care-process contributions & Any such division: the analyte's distribution among the untested is not recoverable, and sentinel codes fix PID & How unmeasured values were encoded & Theorem~1, Corollary~1 \\
Cross-site test of $Y \indep E \mid Z$, failed & That the conjunction of A4, A6a, and A6b fails & Which assumption failed; that biology differs & Sites, measurements, representation; effect sizes with uncertainty & Theorems~2 and~3 \\
Same test, passed & Corroboration of the conjunction & That A4 holds, even given A6 & That the joint hypothesis, not A4 alone, was tested & Proposition~1 \\
Adding ordering indicators and values & Sharper posteriors if the record pins down relevant physiology & Identification in general & Whether ordering practice differs by site & SI Corollary~2 \\
Sensitivity analysis over case mix and representation loss & Whether A4 is compatible with the observed gap for posited error rates & The error rates, except conditionally on A4 and a parametric latent model & The sensitivity model and the error rate needed to explain the gap & Proposition~2, Section~\ref{sec:a-worked-instance} \\
Cross-site comparison using measured tests & Whether A4 holds, given an exact summary of outcome-relevant physiology, or two conditionally independent tests and differing case mix & The sufficiency or conditional-independence assumptions themselves & Which tests were used, whether they were routine, and the argument for these assumptions & Propositions~4 and~5, Corollary~3 \\
Interventional or quasi-experimental variation in ordering & Potentially, mechanism-level contrasts & Outside the scope of these theorems & Whether such variation was available and used & Section~\ref{sec:what-remains-actionable} \\
\bottomrule
\end{tabularx}
\end{table*}
\subsection{Implications for representation learning, invariant prediction, and equity}\label{sec:implications}

A representation may transfer well across sites because mechanisms are shared or because site-specific mechanisms and case mix cancel within representation cells (Proposition~1); it may transfer poorly because mechanisms differ or because identical mechanisms are seen through a lossy representation of different populations (Proposition~2). External validation therefore estimates performance in a new setting, which is what it is for; it does not license a statement about mechanism. Methods that seek predictors stable across environments \cite{peters,arjovsky,rothenhausler} likewise need representation sufficiency (Assumption~6b) as a stated premise when invariance is sought in a learned representation rather than in the latent state.

The construction also bears on health equity \cite{obermeyer,vyas}. When populations differ in both case mix and care, a difference produced by unequal care can be read as a biological difference between populations, which is one way institutional inequity comes to be encoded as apparent biology \cite{vyas}. The reverse error is also possible: a real physiological difference can be misread as a practice effect. Claims that attribute group or site differences to biology need justification from outside the observational data.

\subsection{What remains actionable}\label{sec:what-remains-actionable}

Four forms of evidence survive the result. \emph{Falsification and corroboration:} repeated survival of the invariance test across independent site pairs and laboratory types corroborates the joint hypothesis, and a reliable failure falsifies it without localizing which part failed. \emph{Sensitivity analysis:} Eqs.~\eqref{eq:7} and~\eqref{eq:8} give the cross-site gap that biological invariance alone would produce for posited case-mix contrasts and error rates. This turns an unanswerable yes-or-no question into a bounded, reportable quantity, in the spirit of partial identification \cite{manski90,manski03}. \emph{Routine-measurement anchoring:} protocolized, near-universally ordered measurements remove the ordering gate (Section~\ref{sec:routine}). On their own they remove only the site-specific ordering pathway of SI Corollary~2, not the case-mix pathway of Proposition~2; an exact summary of the outcome-relevant physiology, or two tests with independent errors at sites with different case mix, make the invariance comparison an identification strategy (Propositions~4 and~5), and a routine reference standard turns the sensitivity analysis into a calibration (Corollary~3). \emph{Additional identifying variation:} randomized or protocol-driven changes in ordering practice, such as stepped-wedge guideline changes or natural experiments, supply information absent from the observational distribution. Digital twins \cite{corralacero}, which couple EHR data with mechanistic physiological models, and retrieval-augmented language models \cite{lewis,zakka} applied to each site's written ordering protocols could bring in information of a related kind: a mechanistic model supplies an explicit assumption about \(Z^{\star}\), and documented protocols constrain the ordering process directly. Neither removes the boundary, and neither has been shown to identify the mechanism, but both turn implicit assumptions into stated, checkable ones.

The title's metaphor is meant in this sense. Dark matter is inferred from its gravitational effects on what can be observed, and its properties are constrained by triangulating across independent channels. Clinical judgment stands in a similar relation to the EHR: ordering intent and the practice component of the outcome mechanism are never recorded, yet they shape which values exist (Theorem~1) and how representation cells relate to outcomes (Theorem~3). Bell's theorem \cite{bell} is a point of comparison for the logical form of the result, though only for the form: Bell showed that every local hidden-variable model implies an inequality on measurable correlations, so a hypothesis about unobservables can be tested. Our results concern the same kind of question, which latent-variable models are compatible with which observed distributions, and land on the other side of it: biological invariance implies no observable constraint on its own (SI Remark~1), and acquires one only when conjoined with Assumption~6 (Theorem~2). The practical consequence is that clinical judgment has to be constrained by triangulation across sites, measurement types, and interventional variation; a single observational dataset cannot model it away.

\section{Limitations}\label{sec:limitations}

Theorem~1 and Corollary~1 hold for any joint law given Assumptions~2 and 3; part~(i) of Theorem~1 assumes full support among tested patients and no restriction on how ordering depends on the analyte beyond Assumptions~2 and~6a, and under the recoverability conditions of the missingness-graph literature \cite{mohanpearl2021} parts of the joint law become identified; part~(iii) additionally requires re-encoding invariance of the redundancy measure. Propositions~4 and~5 exchange the observational limit for assumptions the data cannot check, an exact summary of the outcome-relevant state or conditionally independent tests of binary physiology. The other results are exact within the stated model class, except that Proposition~3 is generic rather than universal. The model class holds the representation channel fixed across sites and builds in exclusion (Assumption~6a). Relaxing the first only strengthens non-identifiability. The second is substantive: where a test result triggers treatment that changes prognosis independently of physiology, the outcome mechanism must include \(S\), and the identification problem changes form; the same applies when \(Y\) is mortality and early deaths are not separated by a landmark design (Section~\ref{sec:assumptions}). The worked example and the sensitivity calculation are binary, with a symmetric, site-invariant channel. Lemma~1 and Theorem~3 do not require discreteness, but a continuous analogue of Proposition~3 and general sensitivity models remain open. The framework is cross-sectional and treats the encoder as fixed. The invariance test of Section~\ref{sec:cross-site-invariance-is-a-falsifiable-implication} is stated for discrete \(Z\); with a continuous embedding, binning or a conditional-independence test introduces choices that we do not analyze. Theorem~3 concerns point identification of a binary property and does not rule out informative partial identification of the degree of invariance violation \cite{manski90,manski03}.

\section{Conclusion}\label{sec:conclusion}

Within the model class studied here, biological invariance across clinical environments is not identifiable from observational data. Whether representation-conditional outcomes differ across sites (Theorem~3) or agree (Proposition~1, for a binary outcome with cell rates strictly between 0 and 1; SI Remark~1), the observed distribution is generated both by a model in which the biological mechanism is site-invariant and by one in which it is not. Because the gap lies in the relation between the data-generating process and the data rather than in sampling error, more data and better estimators leave it where it is. Observational EHR research remains informative. Cross-site invariance is a falsifiable implication of a clearly stated joint hypothesis. Sensitivity analysis can bound how much of an observed difference case mix and representation loss could explain, and interventional variation can supply what passive observation cannot. The coupling of patient and care in routine clinical data is therefore an identification boundary; no estimator applied to the observational distribution alone removes it. What we recommend is to state the boundary and report what can be learned inside it, and to look outside it for identifying variation.

\section*{Declarations}

\paragraph{Acknowledgments.} We thank Drago Ple\v{c}ko for a detailed review of an earlier version of this manuscript, which led to Section~\ref{sec:layers}, Theorem~1(i), and Section~\ref{sec:routine}.

\paragraph{Funding.} DR is supported by the European Union's Horizon Europe research and innovation program under the Marie Sk\l{}odowska-Curie COFUND grant agreement No.~101127936 (DeMythif.AI). BFA is supported by a fellowship from “la Caixa” Foundation (ID 100010434), fellowship code B006351. 

\paragraph{Competing interests.} The authors declare no competing interests.

\paragraph{Ethics approval.} Not applicable. This is a theoretical study; no human participants, patient records or other personal data were used.

\paragraph{Use of AI tools.} The authors used large language models during manuscript preparation: a model developed internally by the authors' research group for language editing, and Claude (Anthropic) to check proofs and arithmetic, verify references, edit language, and draft revisions of selected passages and proofs, including parts of Sections~\ref{sec:layers}, \ref{sec:single-dataset-decomposition-is-ill-posed}, and~\ref{sec:routine}, the sensitivity analysis in Section~\ref{sec:a-worked-instance}, and SI~S3 to S5, in response to the authors' and co-authors' review. The study's conception, results and conclusions are the authors'. All AI-assisted text was reviewed, checked and approved by the authors, who take full responsibility for the content of the manuscript.

\paragraph{Data and code availability.} No patient data were used. Every number in the worked example and sensitivity analysis can be reproduced by hand from SI~S5; Python scripts that reproduce them and check Proposition~5 (\texttt{reproduce\_example.py}, \texttt{check\_two\_tests.py}) accompany this article.

\paragraph{ORCID iDs.} M.~Shahin 0009-0006-7717-5310; K.~Chaisutyakorn 0009-0006-7395-6830; D.~Proios 0000-0003-1089-8726; V.~Mishra 0000-0002-4149-7119; B.~Zhang 0000-0002-4439-8212; A.~Sivaraman 0009-0009-0156-3546; S.~Roy 0009-0004-5965-9368; B.~Fit\'{e} Abril 0009-0009-9184-8579; T.~Y.-T.~Chen 0000-0002-1368-8161; H.~G.~Kwak 0000-0002-0847-5073; V.~Mathur 0009-0001-2571-6406; L.~A.~Celi 0000-0001-6712-6626.

\bibliographystyle{unsrtnat}
\bibliography{references}

\begin{thebibliography}{46}
\providecommand{\natexlab}[1]{#1}
\providecommand{\url}[1]{\texttt{#1}}
\expandafter\ifx\csname urlstyle\endcsname\relax
  \providecommand{\doi}[1]{doi: #1}\else
  \providecommand{\doi}{doi: \begingroup \urlstyle{rm}\Url}\fi

\bibitem[Agniel et~al.(2018)Agniel, Kohane, and Weber]{agniel}
Denis Agniel, Isaac~S. Kohane, and Griffin~M. Weber.
\newblock Biases in electronic health record data due to processes within the
  healthcare system: retrospective observational study.
\newblock \emph{BMJ}, 361:\penalty0 k1479, 2018.
\newblock \doi{10.1136/bmj.k1479}.

\bibitem[Pivovarov et~al.(2014)Pivovarov, Albers, Sepulveda, and
  Elhadad]{pivovarov}
Rimma Pivovarov, David~J. Albers, Jorge~L. Sepulveda, and No{\'e}mie Elhadad.
\newblock Identifying and mitigating biases in {EHR} laboratory tests.
\newblock \emph{Journal of Biomedical Informatics}, 51:\penalty0 24--34, 2014.
\newblock \doi{10.1016/j.jbi.2014.03.016}.

\bibitem[Hripcsak and Albers(2013)]{hripcsak}
George Hripcsak and David~J. Albers.
\newblock Next-generation phenotyping of electronic health records.
\newblock \emph{Journal of the American Medical Informatics Association},
  20\penalty0 (1):\penalty0 117--121, 2013.
\newblock \doi{10.1136/amiajnl-2012-001145}.

\bibitem[Groenwold(2020)]{groenwold}
Rolf H.~H. Groenwold.
\newblock Informative missingness in electronic health record systems: the
  curse of knowing.
\newblock \emph{Diagnostic and Prognostic Research}, 4\penalty0 (1):\penalty0
  8, 2020.
\newblock \doi{10.1186/s41512-020-00077-0}.

\bibitem[Finlayson et~al.(2021)Finlayson, Subbaswamy, Singh, Bowers, Kupke,
  Zittrain, et~al.]{finlayson}
Samuel~G. Finlayson, Adarsh Subbaswamy, Karandeep Singh, John Bowers, Annabel
  Kupke, Jonathan Zittrain, et~al.
\newblock The clinician and dataset shift in artificial intelligence.
\newblock \emph{New England Journal of Medicine}, 385\penalty0 (3):\penalty0
  283--286, 2021.
\newblock \doi{10.1056/NEJMc2104626}.
\newblock Letter.

\bibitem[Phelan et~al.(2017)Phelan, Bhavsar, and Goldstein]{phelan}
Matthew Phelan, Nrupen~A. Bhavsar, and Benjamin~A. Goldstein.
\newblock Illustrating informed presence bias in electronic health records
  data: how patient interactions with a health system can impact inference.
\newblock \emph{eGEMs (Generating Evidence \& Methods to improve patient
  outcomes)}, 5\penalty0 (1):\penalty0 22, 2017.
\newblock \doi{10.5334/egems.243}.

\bibitem[Goldstein et~al.(2016)Goldstein, Bhavsar, Phelan, and
  Pencina]{goldstein}
Benjamin~A. Goldstein, Nrupen~A. Bhavsar, Matthew Phelan, and Michael~J.
  Pencina.
\newblock Controlling for informed presence bias due to the number of health
  encounters in an electronic health record.
\newblock \emph{American Journal of Epidemiology}, 184\penalty0 (11):\penalty0
  847--855, 2016.
\newblock \doi{10.1093/aje/kww112}.

\bibitem[Sauer et~al.(2022)Sauer, Chen, Hyland, Girbes, Elbers, and
  Celi]{sauer}
Christopher~M. Sauer, Li-Ching Chen, Stephanie~L. Hyland, Armand Girbes, Paul
  Elbers, and Leo~A. Celi.
\newblock Leveraging electronic health records for data science: common
  pitfalls and how to avoid them.
\newblock \emph{The Lancet Digital Health}, 4\penalty0 (12):\penalty0
  e893--e898, 2022.
\newblock \doi{10.1016/S2589-7500(22)00154-6}.

\bibitem[Rossetti et~al.(2025)Rossetti, Dykes, Knaplund, Cho, Withall,
  Lowenthal, et~al.]{rossetti}
Sarah~C. Rossetti, Patricia~C. Dykes, Chris Knaplund, Sandy Cho, Jennifer
  Withall, Graham Lowenthal, et~al.
\newblock Real-time surveillance system for patient deterioration: a pragmatic
  cluster-randomized controlled trial.
\newblock \emph{Nature Medicine}, 31\penalty0 (6):\penalty0 1895--1902, 2025.
\newblock \doi{10.1038/s41591-025-03609-7}.

\bibitem[Obermeyer et~al.(2019)Obermeyer, Powers, Vogeli, and
  Mullainathan]{obermeyer}
Ziad Obermeyer, Brian Powers, Christine Vogeli, and Sendhil Mullainathan.
\newblock Dissecting racial bias in an algorithm used to manage the health of
  populations.
\newblock \emph{Science}, 366\penalty0 (6464):\penalty0 447--453, 2019.
\newblock \doi{10.1126/science.aax2342}.

\bibitem[Vyas et~al.(2020)Vyas, Eisenstein, and Jones]{vyas}
Darshali~A. Vyas, Leo~G. Eisenstein, and David~S. Jones.
\newblock Hidden in plain sight: reconsidering the use of race correction in
  clinical algorithms.
\newblock \emph{New England Journal of Medicine}, 383\penalty0 (9):\penalty0
  874--882, 2020.
\newblock \doi{10.1056/NEJMms2004740}.

\bibitem[Gianfrancesco et~al.(2018)Gianfrancesco, Tamang, Yazdany, and
  Schmajuk]{gianfrancesco}
Milena~A. Gianfrancesco, Suzanne Tamang, Jinoos Yazdany, and Gabriela Schmajuk.
\newblock Potential biases in machine learning algorithms using electronic
  health record data.
\newblock \emph{JAMA Internal Medicine}, 178\penalty0 (11):\penalty0
  1544--1547, 2018.
\newblock \doi{10.1001/jamainternmed.2018.3763}.

\bibitem[Rubin(1976)]{rubin}
Donald~B. Rubin.
\newblock Inference and missing data.
\newblock \emph{Biometrika}, 63\penalty0 (3):\penalty0 581--592, 1976.
\newblock \doi{10.1093/biomet/63.3.581}.

\bibitem[Kauffman et~al.(2025)Kauffman, Miotto, Klang, Costa, Norgeot, Zitnik,
  et~al.]{kauffman}
Justin Kauffman, Riccardo Miotto, Eyal Klang, Anthony Costa, Beau Norgeot,
  Marinka Zitnik, et~al.
\newblock Embedding methods for electronic health record research.
\newblock \emph{Annual Review of Biomedical Data Science}, 8\penalty0
  (1):\penalty0 563--590, 2025.
\newblock \doi{10.1146/annurev-biodatasci-103123-094729}.

\bibitem[Subbaswamy and Saria(2020)]{subbaswamy}
Adarsh Subbaswamy and Suchi Saria.
\newblock From development to deployment: dataset shift, causality, and
  shift-stable models in health {AI}.
\newblock \emph{Biostatistics}, 21\penalty0 (2):\penalty0 345--352, 2020.
\newblock \doi{10.1093/biostatistics/kxz041}.

\bibitem[Peters et~al.(2016)Peters, B{\"u}hlmann, and Meinshausen]{peters}
Jonas Peters, Peter B{\"u}hlmann, and Nicolai Meinshausen.
\newblock Causal inference by using invariant prediction: identification and
  confidence intervals.
\newblock \emph{Journal of the Royal Statistical Society: Series B (Statistical
  Methodology)}, 78\penalty0 (5):\penalty0 947--1012, 2016.
\newblock \doi{10.1111/rssb.12167}.

\bibitem[Arjovsky et~al.(2019)Arjovsky, Bottou, Gulrajani, and
  Lopez-Paz]{arjovsky}
Martin Arjovsky, L{\'e}on Bottou, Ishaan Gulrajani, and David Lopez-Paz.
\newblock Invariant risk minimization.
\newblock arXiv:1907.02893, 2019.

\bibitem[Rothenh{\"a}usler et~al.(2021)Rothenh{\"a}usler, Meinshausen,
  B{\"u}hlmann, and Peters]{rothenhausler}
Dominik Rothenh{\"a}usler, Nicolai Meinshausen, Peter B{\"u}hlmann, and Jonas
  Peters.
\newblock Anchor regression: heterogeneous data meet causality.
\newblock \emph{Journal of the Royal Statistical Society: Series B (Statistical
  Methodology)}, 83\penalty0 (2):\penalty0 215--246, 2021.
\newblock \doi{10.1111/rssb.12398}.

\bibitem[Zhang et~al.(2013)Zhang, Sch{\"o}lkopf, Muandet, and Wang]{zhang}
Kun Zhang, Bernhard Sch{\"o}lkopf, Krikamol Muandet, and Zhikun Wang.
\newblock Domain adaptation under target and conditional shift.
\newblock In \emph{Proceedings of the 30th International Conference on Machine
  Learning}, volume~28 of \emph{Proceedings of Machine Learning Research},
  pages 819--827. PMLR, 2013.

\bibitem[Pearl(2009{\natexlab{a}})]{pearl}
Judea Pearl.
\newblock \emph{Causality: Models, Reasoning, and Inference}.
\newblock Cambridge University Press, Cambridge, 2nd edition,
  2009{\natexlab{a}}.

\bibitem[Pearl and Bareinboim(2014)]{pearlbareinboim}
Judea Pearl and Elias Bareinboim.
\newblock External validity: from do-calculus to transportability across
  populations.
\newblock \emph{Statistical Science}, 29\penalty0 (4):\penalty0 579--595, 2014.
\newblock \doi{10.1214/14-STS486}.

\bibitem[Bareinboim and Pearl(2016)]{bareinboim}
Elias Bareinboim and Judea Pearl.
\newblock Causal inference and the data-fusion problem.
\newblock \emph{Proceedings of the National Academy of Sciences}, 113\penalty0
  (27):\penalty0 7345--7352, 2016.
\newblock \doi{10.1073/pnas.1510507113}.

\bibitem[Manski(1990)]{manski90}
Charles~F. Manski.
\newblock Nonparametric bounds on treatment effects.
\newblock \emph{American Economic Review}, 80\penalty0 (2):\penalty0 319--323,
  1990.

\bibitem[Manski(2003)]{manski03}
Charles~F. Manski.
\newblock \emph{Partial Identification of Probability Distributions}.
\newblock Springer, New York, 2003.

\bibitem[Williams and Beer(2010)]{wb}
Paul~L. Williams and Randall~D. Beer.
\newblock Nonnegative decomposition of multivariate information.
\newblock arXiv:1004.2515, 2010.

\bibitem[Bertschinger et~al.(2014)Bertschinger, Rauh, Olbrich, Jost, and
  Ay]{bertschinger}
Nils Bertschinger, Johannes Rauh, Eckehard Olbrich, J{\"u}rgen Jost, and Nihat
  Ay.
\newblock Quantifying unique information.
\newblock \emph{Entropy}, 16\penalty0 (4):\penalty0 2161--2183, 2014.
\newblock \doi{10.3390/e16042161}.

\bibitem[Griffith and Koch(2014)]{griffith}
Virgil Griffith and Christof Koch.
\newblock Quantifying synergistic mutual information.
\newblock In Mikhail Prokopenko, editor, \emph{Guided Self-Organization:
  Inception}, volume~9 of \emph{Emergence, Complexity and Computation}, pages
  159--190. Springer, Berlin, Heidelberg, 2014.
\newblock \doi{10.1007/978-3-642-53734-9_6}.

\bibitem[Choi et~al.(2026)Choi, Jeanselme, Elias, and Joshi]{choi}
Young~Sang Choi, Vincent Jeanselme, Pierre Elias, and Shalmali Joshi.
\newblock {ICYM$^2$I}: The illusion of multimodal informativeness under
  missingness.
\newblock In \emph{International Conference on Learning Representations
  (ICLR)}, 2026.
\newblock arXiv:2505.16953.

\bibitem[VanderWeele et~al.(2012)VanderWeele, Valeri, and
  Ogburn]{vanderweele2012}
Tyler~J. VanderWeele, Linda Valeri, and Elizabeth~L. Ogburn.
\newblock The role of measurement error and misclassification in mediation
  analysis.
\newblock \emph{Epidemiology}, 23\penalty0 (4):\penalty0 561--564, 2012.
\newblock \doi{10.1097/EDE.0b013e318258f5e4}.

\bibitem[le~Cessie et~al.(2012)le~Cessie, Debeij, Rosendaal, Cannegieter, and
  Vandenbroucke]{lecessie}
Saskia le~Cessie, Jan Debeij, Frits~R. Rosendaal, Suzanne~C. Cannegieter, and
  Jan~P. Vandenbroucke.
\newblock Quantification of bias in direct effects estimates due to different
  types of measurement error in the mediator.
\newblock \emph{Epidemiology}, 23\penalty0 (4):\penalty0 551--560, 2012.
\newblock \doi{10.1097/EDE.0b013e318254f5de}.

\bibitem[Kuroki and Pearl(2014)]{kurokipearl}
Manabu Kuroki and Judea Pearl.
\newblock Measurement bias and effect restoration in causal inference.
\newblock \emph{Biometrika}, 101\penalty0 (2):\penalty0 423--437, 2014.
\newblock \doi{10.1093/biomet/ast066}.

\bibitem[Miao et~al.(2018)Miao, Geng, and Tchetgen~Tchetgen]{miao}
Wang Miao, Zhi Geng, and Eric~J. Tchetgen~Tchetgen.
\newblock Identifying causal effects with proxy variables of an unmeasured
  confounder.
\newblock \emph{Biometrika}, 105\penalty0 (4):\penalty0 987--993, 2018.
\newblock \doi{10.1093/biomet/asy038}.

\bibitem[Allman et~al.(2009)Allman, Matias, and Rhodes]{allman}
Elizabeth~S. Allman, Catherine Matias, and John~A. Rhodes.
\newblock Identifiability of parameters in latent structure models with many
  observed variables.
\newblock \emph{The Annals of Statistics}, 37\penalty0 (6A):\penalty0
  3099--3132, 2009.
\newblock \doi{10.1214/09-AOS689}.

\bibitem[Hui and Walter(1980)]{huiwalter}
Siu~L. Hui and Stephen~D. Walter.
\newblock Estimating the error rates of diagnostic tests.
\newblock \emph{Biometrics}, 36\penalty0 (1):\penalty0 167--171, 1980.
\newblock \doi{10.2307/2530508}.

\bibitem[Zhang and Rubin(2003)]{zhangrubin}
Junni~L. Zhang and Donald~B. Rubin.
\newblock Estimation of causal effects via principal stratification when some
  outcomes are truncated by ``death''.
\newblock \emph{Journal of Educational and Behavioral Statistics}, 28\penalty0
  (4):\penalty0 353--368, 2003.
\newblock \doi{10.3102/10769986028004353}.

\bibitem[Mohan et~al.(2013)Mohan, Pearl, and Tian]{mohan2013}
Karthika Mohan, Judea Pearl, and Jin Tian.
\newblock Graphical models for inference with missing data.
\newblock In \emph{Advances in Neural Information Processing Systems 26 (NIPS
  2013)}, pages 1277--1285, 2013.

\bibitem[Mohan and Pearl(2021)]{mohanpearl2021}
Karthika Mohan and Judea Pearl.
\newblock Graphical models for processing missing data.
\newblock \emph{Journal of the American Statistical Association}, 116\penalty0
  (534):\penalty0 1023--1037, 2021.
\newblock \doi{10.1080/01621459.2021.1874961}.

\bibitem[Pearl(2009{\natexlab{b}})]{pearl2009overview}
Judea Pearl.
\newblock Causal inference in statistics: an overview.
\newblock \emph{Statistics Surveys}, 3:\penalty0 96--146, 2009{\natexlab{b}}.
\newblock \doi{10.1214/09-SS057}.

\bibitem[Bareinboim et~al.(2022)Bareinboim, Correa, Ibeling, and
  Icard]{bareinboim2022}
Elias Bareinboim, Juan~D. Correa, Duligur Ibeling, and Thomas Icard.
\newblock On {P}earl's hierarchy and the foundations of causal inference.
\newblock In \emph{Probabilistic and Causal Inference: The Works of {J}udea
  {P}earl}, pages 507--556. ACM, New York, NY, USA, 2022.
\newblock \doi{10.1145/3501714.3501743}.

\bibitem[Sperrin et~al.(2020)Sperrin, Martin, Sisk, and Peek]{sperrin}
Matthew Sperrin, Glen~P. Martin, Rose Sisk, and Niels Peek.
\newblock Missing data should be handled differently for prediction than for
  description or causal explanation.
\newblock \emph{Journal of Clinical Epidemiology}, 125:\penalty0 183--187,
  2020.
\newblock \doi{10.1016/j.jclinepi.2020.03.028}.

\bibitem[Meek(1995)]{meek}
Christopher Meek.
\newblock Strong completeness and faithfulness in {B}ayesian networks.
\newblock In \emph{Proceedings of the Eleventh Conference on Uncertainty in
  Artificial Intelligence (UAI)}, pages 411--418. Morgan Kaufmann, 1995.

\bibitem[Spirtes et~al.(2000)Spirtes, Glymour, and Scheines]{sgs}
Peter Spirtes, Clark Glymour, and Richard Scheines.
\newblock \emph{Causation, Prediction, and Search}.
\newblock MIT Press, Cambridge, MA, 2nd edition, 2000.

\bibitem[Corral-Acero et~al.(2020)Corral-Acero, Margara, Marciniak, Rodero,
  Loncaric, Feng, et~al.]{corralacero}
Jorge Corral-Acero, Francesca Margara, Maciej Marciniak, Cristobal Rodero,
  Filip Loncaric, Yingjing Feng, et~al.
\newblock The `{D}igital {T}win' to enable the vision of precision cardiology.
\newblock \emph{European Heart Journal}, 41\penalty0 (48):\penalty0 4556--4564,
  2020.
\newblock \doi{10.1093/eurheartj/ehaa159}.

\bibitem[Lewis et~al.(2020)Lewis, Perez, Piktus, Petroni, Karpukhin, Goyal,
  et~al.]{lewis}
Patrick Lewis, Ethan Perez, Aleksandra Piktus, Fabio Petroni, Vladimir
  Karpukhin, Naman Goyal, et~al.
\newblock Retrieval-augmented generation for knowledge-intensive {NLP} tasks.
\newblock In \emph{Advances in Neural Information Processing Systems},
  volume~33, pages 9459--9474, 2020.

\bibitem[Zakka et~al.(2024)Zakka, Shad, Chaurasia, Dalal, Kim, Moor,
  et~al.]{zakka}
Cyril Zakka, Rohan Shad, Akash Chaurasia, Alex~R. Dalal, Jennifer~L. Kim,
  Michael Moor, et~al.
\newblock Almanac---retrieval-augmented language models for clinical medicine.
\newblock \emph{NEJM AI}, 1\penalty0 (2):\penalty0 AIoa2300068, 2024.
\newblock \doi{10.1056/AIoa2300068}.

\bibitem[Bell(1964)]{bell}
John~S. Bell.
\newblock On the {E}instein {P}odolsky {R}osen paradox.
\newblock \emph{Physics Physique Fizika}, 1\penalty0 (3):\penalty0 195--200,
  1964.
\newblock \doi{10.1103/PhysicsPhysiqueFizika.1.195}.

\end{thebibliography}

\end{document}